\documentclass{article}

\usepackage[final, nohyperref]{corl_2020} 

\usepackage{graphicx} 
\usepackage{wrapfig}
\usepackage{amsmath} 
\usepackage{subcaption}

\title{\LARGE \bf
Estimating Mass Distribution of Articulated Objects using Non-prehensile Manipulation
}

\author{K. Niranjan Kumar \\
Georgia Institute of Technology \\
\texttt{niranjankumar@gatech.edu}
\And
Irfan Essa \\
Georgia Institute of Technology \\
\texttt{irfan@gatech.edu}
\And
Sehoon Ha \\
Georgia Institute of Technology \\
\texttt{sehoonha@gatech.edu}
\And
C. Karen Liu \\
Georgia Institute of Technology \\ Stanford University \\
\texttt{karenliu@cs.stanford.edu}
}

\begin{document}

\maketitle
\thispagestyle{empty}
\pagestyle{empty}



\newcommand{\old}[1]{\textcolor{red}{\textbf {\st{#1}}}}
\newcommand{\new}[1]{\textcolor{red}{#1}}
\newcommand{\note}[1]{\cmt{Note: #1}}
\newcommand{\karen}[1]{\textcolor{red}{{Karen: #1}}}
\newcommand{\sehoon}[1]{\textcolor{magenta}{{Sehoon: #1}}}

\newcommand{\friedl}[1]{\textcolor{orange}{{Friedl: #1}}}
\newcommand{\tom}[1]{\textcolor{orange}{{Tom: #1}}}
\newcommand{\ck}[1]{\textcolor{Plum}{{Charlie: #1}}}
\newcommand{\original}[1]{\textcolor{blue}{{Original: #1}}}
\newcommand{\jie}[1]{\textcolor{blue}{{Jie: #1}}}
\newcommand{\alex}[1]{\textcolor{Purple}{{Alex: #1}}}
\newcommand{\greg}[1]{\textcolor{green}{{Greg: #1}}}
\newcommand{\zackory}[1]{\textcolor{Orange}{{Zackory: #1}}}
\newcommand{\charlie}[1]{\textcolor{cyan}{{Charlie: #1}}}
\newcommand{\henry}[1]{\textcolor{TealBlue}{{Henry: #1}}}
\newcommand{\wenhao}[1]
{\textcolor{ForestGreen}{{Wenhao: #1}}}
\newcommand{\yifeng}[1]
{\textcolor{darkblue}{{Yifeng: #1}}}
\newcommand{\newtext}[1]{#1}
\newcommand{\eqnref}[1]{Equation~(\ref{eqn:#1})}

\long\def\ignorethis#1{}

\newcommand{\etal}{{\em{et~al.}\ }}
\newcommand{\eg}{e.g.\ }
\newcommand{\ie}{i.e.\ }

\newcommand{\figtodo}[1]{\framebox[0.8\columnwidth]{\rule{0pt}{1in}#1}}
\newcommand{\figref}[1]{Figure~\ref{fig:#1}}
\newcommand{\secref}[1]{Section~\ref{sec:#1}}

\newcommand{\vc}[1]{\ensuremath{\boldsymbol{#1}}}
\newcommand{\pd}[2]{\ensuremath{\frac{\partial{#1}}{\partial{#2}}}}
\newcommand{\pdd}[3]{\ensuremath{\frac{\partial^2{#1}}{\partial{#2}\,\partial{#3}}}}

\newcommand{\vEndEff}{\ensuremath{\vc{d}}}
\newcommand{\vRelMove}{\ensuremath{\vc{r}}}
\newcommand{\sSet}{\ensuremath{S}}

\newcommand{\vControl}{\ensuremath{\vc{u}}}
\newcommand{\vPoint}{\ensuremath{\vc{p}}}
\newcommand{\sSpringCoef}{{\ensuremath{k_{s}}}}
\newcommand{\sDamperCoef}{{\ensuremath{k_{d}}}}
\newcommand{\vHandle}{\ensuremath{\vc{h}}}
\newcommand{\vForce}{\ensuremath{\vc{f}}}

\newcommand{\mTransChain}{\ensuremath{\vc{W}}}
\newcommand{\mRotateTrans}{\ensuremath{\vc{R}}}
\newcommand{\sJoint}{\ensuremath{q}}
\newcommand{\vJoint}{\ensuremath{\vc{q}}}
\newcommand{\mJoint}{\ensuremath{\vc{Q}}}
\newcommand{\mMass}{\ensuremath{\vc{M}}}
\newcommand{\sMass}{\ensuremath{{m}}}
\newcommand{\vGravity}{\ensuremath{\vc{g}}}
\newcommand{\vConstr}{\ensuremath{\vc{C}}}
\newcommand{\sConstr}{\ensuremath{C}}
\newcommand{\vCOM}{\ensuremath{\vc{x}}}
\newcommand{\sGeneralForce}[1]{\ensuremath{Q_{#1}}}
\newcommand{\vStateVar}{\ensuremath{\vc{y}}}
\newcommand{\vControlVar}{\ensuremath{\vc{u}}}
\newcommand{\argmax}{\operatornamewithlimits{argmax}}
\newcommand{\argmin}{\operatornamewithlimits{argmin}}

\newcommand{\tr}[1]{\ensuremath{\mathrm{tr}\left(#1\right)}}

%
%

\renewcommand{\choose}[2]{\ensuremath{\left(\begin{array}{c} #1 \\ #2 \end{array} \right )}}

\newcommand{\gauss}[3]{\ensuremath{\mathcal{N}(#1 | #2 ; #3)}}

\newcommand{\pctab}{\hspace{0.2in}}
\newenvironment{pseudocode} {\begin{center} \begin{minipage}{\textwidth}
                             \normalsize \vspace{-2\baselineskip} \begin{tabbing}
                             \pctab \= \pctab \= \pctab \= \pctab \=
                             \pctab \= \pctab \= \pctab \= \pctab \= \\}
                            {\end{tabbing} \vspace{-2\baselineskip}
                             \end{minipage} \end{center}}
\newenvironment{items}      {\begin{list}{$\bullet$}
                              {\setlength{\partopsep}{\parskip}
                                \setlength{\parsep}{\parskip}
                                \setlength{\topsep}{0pt}
                                \setlength{\itemsep}{0pt}
                                \settowidth{\labelwidth}{$\bullet$}
                                \setlength{\labelsep}{1ex}
                                \setlength{\leftmargin}{\labelwidth}
                                \addtolength{\leftmargin}{\labelsep}
                                }
                              }
                            {\end{list}}
\newcommand{\newfun}[3]{\noindent\vspace{0pt}\fbox{\begin{minipage}{3.3truein}\vspace{#1}~ {#3}~\vspace{12pt}\end{minipage}}\vspace{#2}}

\newcommand{\key}{\textbf}
\newcommand{\fun}{\textsc}



\begin{abstract}
We explore the problem of estimating the mass distribution of an articulated object by an interactive robotic agent. Our method predicts the mass distribution of an object by using limited sensing and actuating capabilities of a robotic agent that is interacting with the object. We are inspired by the role of exploratory play in human infants. We take the combined approach of supervised and reinforcement learning to train an agent that learns to strategically interact with the object to estimate the object's mass distribution. Our method consists of two neural networks: (i) the policy network which decides how to interact with the object, and (ii) the predictor network that estimates the mass distribution given a history of observations and interactions. Using our method, we train a robotic arm to estimate the mass distribution of an object with moving parts (e.g. an articulated rigid body system) by pushing it on a surface with unknown friction properties. We also demonstrate how our training from simulations can be transferred to real hardware using a small amount of real-world data for fine-tuning. We use a UR10 robot to interact with 3D printed articulated chains with varying mass distributions and show that our method significantly outperforms the baseline system that uses random pushes to interact with the object. Video at \href{https://youtu.be/o3zBdVWvWZw}{https://youtu.be/o3zBdVWvWZw}
\end{abstract}


\section{INTRODUCTION}
Humans continuously make inferences about the physical properties of objects in the world while perceiving and interacting with them~\cite{Spelke2007}.
Relying on our physical intuition, we can predict with a high degree of certainty if a block tower would destabilize and fall~\cite{Hamrick_internalphysics} by simply looking at it. While visual observation is sufficient for an agent to predict a rough outcome of the future, having the ability to physically interact with the surroundings provides more critical information to reason about the world. Infants, for example, engage in exploratory play to discover non-obvious physical properties of objects~\cite{10.2307/1131213}, and develop a core knowledge in long-term cognition about the physical world \cite{10.3389/fpsyg.2018.00635}. Can we teach the robot to engage in a similar exploratory play with a new object in order to learn more about it? 
\begin{figure}[t]
\centering
\includegraphics[width=0.32\textwidth]{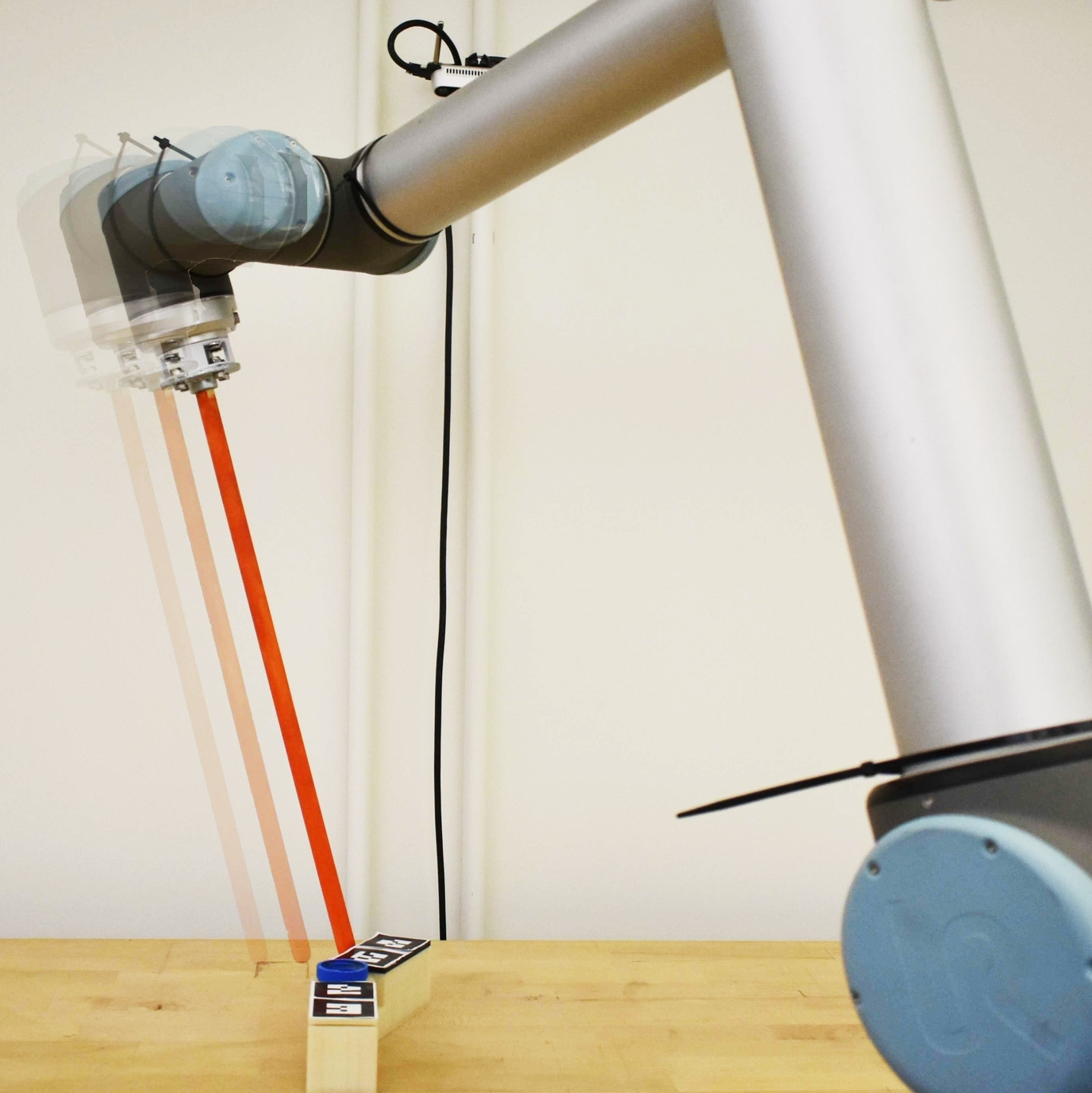}\hfill
\includegraphics[width=0.32\textwidth]{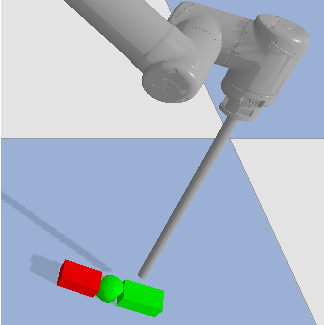}\hfill
\includegraphics[width=0.32\textwidth]{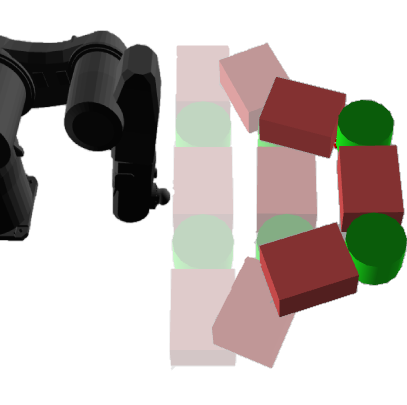}\hfill
\caption{Example of a push (in two frame time-lapse) that can be executed by the robot on a 2-link chain \textbf{(Left)}. The chain loses contact with the end-effector (painted orange in the picture) and continues sliding on the surface briefly before coming to rest. We run simulations where the agent can interact with articulated objects across two different robot platforms \textbf{(middle, right)}}
\label{fig:real_world}
\end{figure}
In this work, we develop an interactive agent to estimate the mass distribution of an articulated object by pushing it around on a surface. We design a simulated world, shown in Figure \ref{fig:real_world} (Middle, Right), and a real robot setup, shown in Figure \ref{fig:real_world} (Left), where a robotic arm can interact with an articulated chain by pushing it at any point on any segment. 

While estimating mass properties through physical interaction has been previously studied \cite{14,12,24}, our problem is uniquely challenging in three aspects. 
First, unlike estimating the mass of a single rigid body, the Euclidean distance between two equilibrium states of an articulated object is insufficient to infer the mass of each moving segment. Compounding with the discontinuous behaviors due to self-collisions among segments, estimating the mass distribution of an articulated object accurately, requires the robot to push the object at the right spots, with the right force, and make the right conclusions about the outcomes of its interactions. 
Second, to make our method applicable to the real-world scenarios, we do not expect the robot to be able to push or track the state of a moving object, which often requires an accurate motion capture system. Instead, our method is only allowed to use the information acquired when the object comes to a complete stop (an equilibrium state). 
Third, we do not require the agent to grasp the object during any part of the interaction and allow only non-prehensile pushing of the object.

Our problem of system identification can benefit from a policy that generates a strategic sequence of interactions, rather than using random actions. For instance, consider the case where two segments are connected with a hinge joint (similar to the object shown in Figure \ref{fig:real_world} (middle)). If we push one of the links with an arbitrary force, it might cause movements that are either too subtle to accurately measure, or too abrupt to stay within the workspace. In addition, similar forces might result in drastically different equilibrium states of object due to self-collisions, and drastically different forces might result in similar equilibrium states due to $2\pi$ cycle in orientation. Therefore, our system must be able to generate informative actions with proper directions and magnitudes, based on the intermediate guess from the previous interactions. 

To this end, we propose a dual network architecture that consists of  (i) a policy network and (ii) a predictor network working in tandem towards the goal. The predictor network observes the reaction of the object to pushes imparted by the robot, and attempts to predict the mass distribution of the object at the end of the episode. The policy network learns how to push the object so that useful information can be extracted from each push. The predictor network learns from a set of state-action trajectories simulated using articulated objects with various mass distributions and friction coefficients. Once learned, the combination of the policy and the predictor networks can reliably estimate the mass distribution of an unseen object with arbitrary segment shapes and friction coefficient under moderate sensing and actuating errors, using only a few pushes.

Our main contribution is to demonstrate that embodied learning of physical parameters of a system benefits from an intelligent interaction policy. Furthermore, we show that not all actions are created equal when an agent interacts with the intention to discover the physical properties of an object. By comparing to a baseline random interaction policy, we show that our agent learns to exploit the subset of informative actions to significantly improve the accuracy of prediction.

\section{RELATED WORK}

Humans utilize intuitive physics model to reason about the world, and predict its change in the near future \cite{46,47}. Building such a physical model has traditionally been approached as a system identification problem \cite{48}, which has the potential to significantly improve the control policies operated in the real world. Recent advances in deep learning methods advocate a new approach towards predicting the future state of the world directly from the recent observation history~\cite{2,5,6,8,13,27,28,40}. By training a mapping between past states and the future state using simulated sequential data, one can predict whether a block tower is going to fall~\cite{4,22,41}, and the outcome of collision between two rigid bodies~\cite{8}.
The predictive model can also be conditioned on external actions that could influence the next state of the system. This model can be used to find the optimal action such that it results in a state preferential to the task, such as training an agent to play billiards~\cite{21} or pushing objects to their goals~\cite{32,20}.

While directly predicting the next state is a powerful tool, some applications focus on identifying specific physical parameters useful for control algorithm analysis, or developing environment-conditioned control policies \cite{33,34, yang2019single}. For example, the degrees of freedom and the range of motion of an articulated rigid body system can be identified from videos of motion \cite{18,23}, or from a robot actively interacting with the object \cite{15,17,19}.
Our work, like those discussed above, falls under the wide umbrella of Interactive Perception \cite{bohg2017interactive}. While previous approaches that deal with articulated objects have looked into using interactive perception to infer the kinematic structure, our work focuses on the more difficult problem of estimating the mass distribution.

Predicting the mass of an object has been of particular interest in the subarea of robotic manipulation. \cite{26} used a force sensor on two finger tips to estimate the mass and the center of mass of blocks. \cite{14} estimated the mass and friction coefficients from videos. \cite{12} learned latent representations of physical properties that correlate with mass and friction. Our work is closely related to \cite{24}, in which a robotic arm is used to push blocks for estimating their mass and friction coefficients. However, our work goes one step further by predicting the mass distribution of an object consisting \textit{multiple} moving parts. Using a random probing policy as proposed by previous work, like \cite{24}, typically results in poor estimation of the mass distribution. Instead, we train a policy that strategically pushes the object to extract maximum information for estimating its mass distribution.

\begin{figure*}[t]
\centerline{\includegraphics[width=0.8\textwidth]{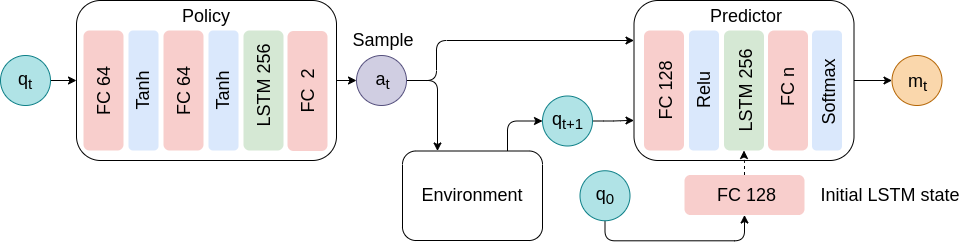}}
\caption{Architecture of our proposed method. The Policy and Predictor networks are trained alternately. During testing, we use the Policy network to suggest actions that are likely to provide novel information about the mass distribution of the object.}

\label{fig:architecture}
\end{figure*}

\section{METHOD}

We introduce a method to enable a robot to estimate the mass distribution of an articulated object by strategically pushing it around on a flat surface. While this problem can be formulated in a variety of ways, our proposed method considers the practical limitations inherent to the robot's sensing and actuation capabilities--it only depends on the information the robot can reliably observe, and the commands the robot can reliably execute. To this end, our method only requires the robot to be able to observe and push the object when it is at an equilibrium state (\ie the object comes to a complete stop), as opposed to transient states where neither accurate state estimation nor precise pushes can be easily achieved. This requirement increases the difficulty of the problem, but makes the deployment of our method to real-world hardware more feasible.

A conventional approach to this problem is to conduct system identification such that the observed data is consistent with the equations of motion. However, since we only observe the object and apply the actions at equilibrium states, we cannot utilize dynamic equations to infer masses in the absence of derivative information (\eg velocity, acceleration). Further, effective system identification also requires analysis on persistent excitation to ensure that the observations span a wide range of system behaviors. Although the object in our problem is a passive multibody system, the mass parameters can only be uniquely identified by a subset of observation and action trajectories (see Supplementary file). This problem is further exacerbated by practical scenarios in which the robot is only expected to interact with the object a few times.

We take a learning approach to circumvent the above challenges. Our method consists of two learned models: a predictor to estimate the mass distribution from observation and action history, and a policy to determine the appropriate push actions to induce useful information for predicting mass distributions. Note that we predict the mass distribution of the object, instead of the actual mass of each segment in the object. Since the kinetic friction force applied on each segment is proportional to the mass of the segment when it lies flat on the surface, the ratio of masses among segments is unaffected by the friction coefficient. Therefore, if we only predict the mass distribution and assume the total mass is known, our method is agnostic to the surface materials of the object.

\subsection{Predicting Mass Distribution from Interaction}

Consider an articulated chain with $n$ rigid bodies of mass connected by revolute joints. Let $\vc{q}$ be the observed configuration of the object, and $\vc{a}$ represent the push the robot exerts on the object. Starting from the initial configuration $\vc{q}_0$, the robot applies a push $\vc{a}_0$ and observes the object when it comes to a complete stop, at $\vc{q}_1$. After repeating this process $K$ times, the robot infers the mass distribution $\vc{m} \in \mathbf{R}^n$ of the object, where each element of $\vc{m}$ indicates the mass of a rigid body normalized by the total mass of the object.

With this problem setting, we learn a predictor $f_\mu$ using supervised learning. The predictor takes as input the belief from the previous step $\vc{h}_t = f_h(\vc{q}_0, \vc{a}_0, \cdots, \vc{a}_{t-1}, \vc{q}_t)$, the current action $\vc{a}_t$, and the next observation at the equilibrium $\vc{q}_{t+1}$, to predict the mass distribution $\vc{m}_t$ at the current step $t$:
\begin{equation}
\label{eq:mapping}
    \vc{m}_t = f_\mu(\vc{h}_t, \vc{a}_t, \vc{q}_{t+1}).
\end{equation}

In practice, we represent $f_\mu$ as a Long Short-Term Memory(LSTM) network (Figure \ref{fig:architecture}) parameterized by $\vc{\mu}$. We define a loss function  as follows:
\begin{equation}\label{eq:loss}
\mathcal{L}(\mu) = \frac{1}{N}\frac{1}{K}\sum_{i=1}^N \sum_{t=0}^{K}{\lVert \vc{m}^i-f_{\mu}(\vc{h}_t, \vc{a}_t^i, \vc{q}_{t+1}^i)\rVert}_2,
\end{equation}
where superscript $i$ is the index of the $N$ training samples. 
Theoretically, we only need the last prediction to be accurate, \ie, $f_\mu(\vc{h}_K, \vc{a}^i_K, \vc{q}^i_{K+1}) = \vc{m}^i$. 
In practice, however, encouraging the earlier prediction ($t < K$) to also be accurate provides a richer reward signal for the reinforcement learning algorithm during policy training.

Minimizing Equation \ref{eq:loss} results in a predictor network that achieves high accuracy when the input trajectory contains sufficient information to identify the mass distribution, \ie, there is a unique mass distribution consistent with the observations induced by given actions. However, as discussed earlier, some input trajectories in Equation \ref{eq:mapping} cannot uniquely identify the mass distribution due to the singularity in the dynamic system, and sparse observations and actions. As a result, the predictor performs poorly on predicting the mass distributions for those regions in the input space. 
\subsection{Learning How to Interact}

This motivates a policy that can generate strategic actions that can uniquely identify the mass distribution.
Given a supervised model that predicts mass distribution from trajectories, what should those trajectories be to get an accurate mass distribution estimate? More specifically, we need to decide what actions to take at each step so that the input trajectory to the predictor lies in the region of high accuracy. We model this problem as a Partially Observed Markov Decision Process (POMDP) where the underlying state $\vc{s_t}$ of the system is defined to be a combination of the pose of the object $\vc{q_t}$, actual system parameters $\bar{\vc{m}}$ and predictor's belief $\vc{h_t}$ of the actual system parameters. The input that is available to the agent, however, is just the pose of the object $\vc{q_t}$ and not the complete state.
Following \cite{wierstra2007solving} we build our policy with an LSTM that can model sufficient statistics $\vc{\hat{h}_t}$ of past inputs. This is because, intuitively, the nature of exploration for our problem requires memorizing how the object has behaved in the past. The objective of our policy is to produce trajectories that can be efficiently reasoned by the predictor to give an accurate estimate of the mass distribution. To this end, we design a reward function that encourages actions that help the predictor achieve a lower error:

\begin{equation}
r( \vc{s}_{t},\vc{a}_t, \vc{s}_{t+1}) = 1- \beta {\|\bar{\vc{m}} - f_\mu(\vc{h}_t, \vc{a}_t, \vc{q}_{t+1}) \|}_1,
\end{equation}
where $\bar{\vc{m}}$ is the ground truth mass distribution and $\beta$ is a multiplicative constant. We design the reward function to be proportional to the negative of the error in mass distribution estimated by the predictor network, and scale its output to be between $[-1,1]$ with $\beta$.
Intuitively, learning a policy can be seen as adapting the input data distribution to operate on low error regions of the predictor. We use a policy gradient method, Proximal Policy Optimization \cite{schulman2017proximal} to solve for the policy $\pi_\theta(\vc{a}|\vc{\hat{h}_t})$, where $\pi$ is represented as a LSTM neural network parameterized by $\theta$ with hidden state $\vc{\hat{h}_t}$ that evolves through time with every new $\vc{q}_t$.

\subsection{Alternating Training Procedure}
In practice, we find that alternating between training the predictor and the policy a few times results in predicting the mass distribution more accurately. We begin by training the predictor on a uniform distribution of action and observation trajectories. Subsequently, we freeze the predictor and train the policy network until convergence. Next, we retrain the predictor with the input trajectories provided by the current policy network. The alternating training procedure terminates when the accuracy of the predictor reaches a target threshold. The training process is terminated with the training of the predictor, as the data distribution imposed on the predictor by the policy might have changed after the previous policy updates. Our alternating scheme gradually refines the training distribution for the predictor from a uniform distribution to a task-relevant one, induced by the policy.

\subsection{Executing Actions} \label{Executing_actions_on_a_robot}
\begin{figure}[t]
\centering
\begin{subfigure}{0.5\textwidth}
\includegraphics[width=0.9\linewidth]{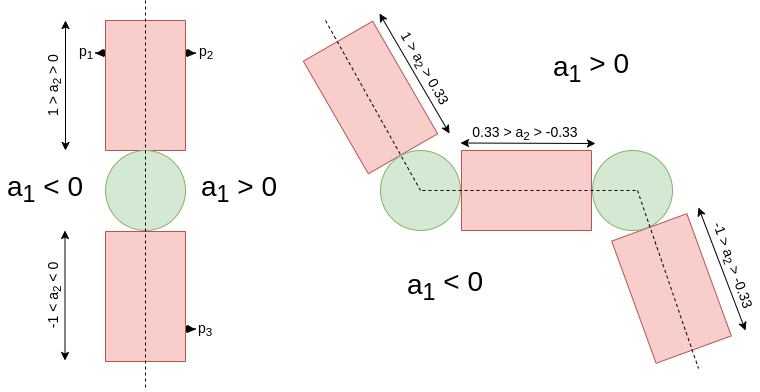}
\caption{}
\label{fig:action_space}
\end{subfigure}%
\begin{subfigure}{0.25\textwidth}
\includegraphics[width=0.95\linewidth]{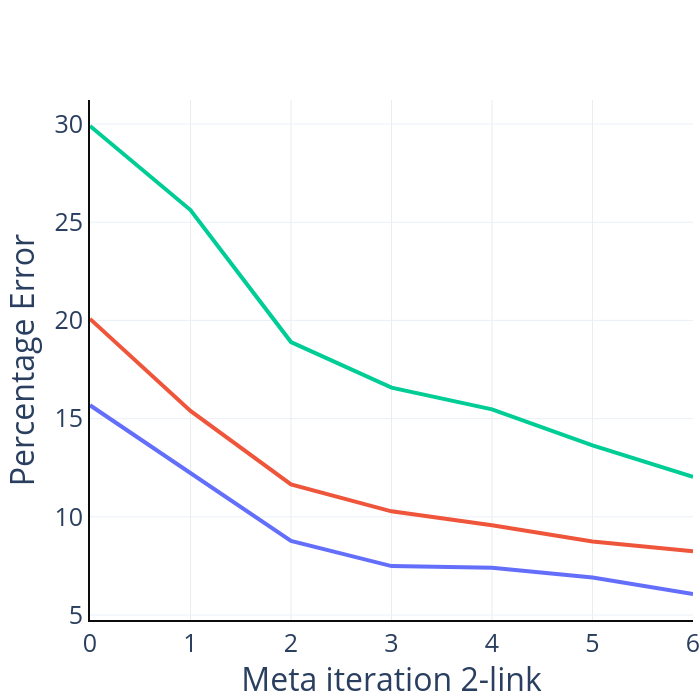}
\caption{}
\label{fig:meta_iter_2link}
\end{subfigure}%
\begin{subfigure}{0.25\textwidth}
\includegraphics[width=0.95\linewidth]{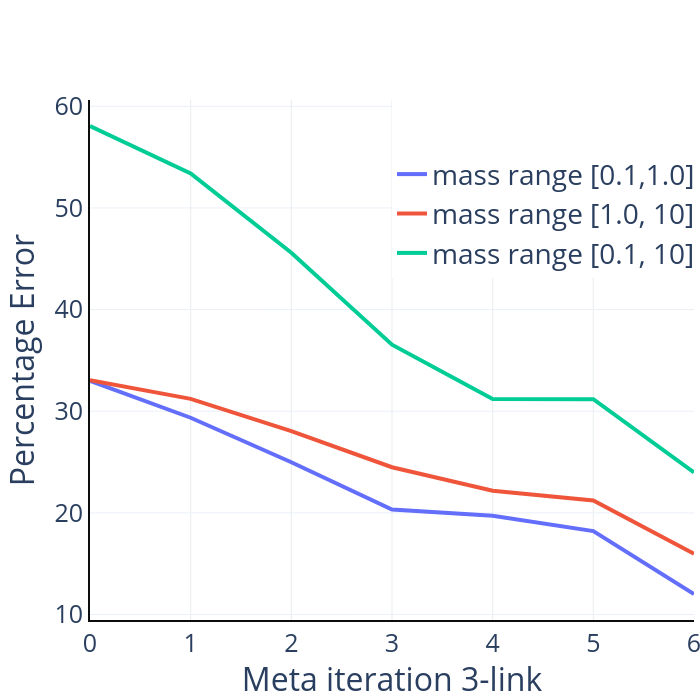}
\caption{}
\label{fig:meta_iter_3link}
\end{subfigure}%
\caption{\textbf{(\ref{fig:action_space})}: We partition a two dimensional action space to represent all permissible pushes that can be executed by the robotic arm. The sign of $a_1$ represents which side of the segment is going to be pushed while $a_2$ spans the points on the selected side. For example all pushes that can be applied to the points $p_1$, $p_2$ and $p_3$ are represented by $\vc{a}=\{-k,0.75\}$, $\vc{a}=\{k,0.75\}$ and $\vc{a}=\{k,-0.75\}$ respectively, where $k$ is a positive real number. \textbf{(\ref{fig:meta_iter_2link})}, \textbf{(\ref{fig:meta_iter_3link})}:  Plot of percentage error vs meta iteration number for two-link and three-link chains respectively.}
\label{fig:action_space_meta_iter}
\end{figure}

We assume that the robot is capable of observing the configuration of the object directly. This assumption can be relaxed by using image input and adding convolution neural networks to the predictor and the policy, without modifying the core of the proposed framework. We further assume that the robot is able to command torques at each actuator and that its base is able to move and rotate on the surface where the object lies. Based on these assumptions, we define the observation vector as the configuration of the object, $\vc{q} = \{x, y, \alpha, \vc{\theta}\}$, where $x$, $y$, $\alpha$ are the global translation and yaw angle of the root link, while $\vc{\theta}$ corresponds to the joint angles of the rest of the chain. The action vector is defined as $\vc{a} = \{a_1, a_2\}$, where $a_1 \in \mathbf{R}$ represents the target velocity of the end-effector when the robot delivers the push, and $a_2\in \mathbf{R}$ is used to select the point of application of the push. Depending on the value of $a_2$ and sign of $a_1$, different points on different chain segments are chosen. We only allow pushes at points that lie on the same horizontal plane as the center of the chain segment and partition the 2D action vector to represent all permissible pushes (Figure \ref{fig:action_space}).

We use the operational space control \cite{khatib1987unified} formalism to generate appropriate joint torques that achieve a desired linear and angular acceleration $\ddot{\vc{x}} \in \mathbf{R}^6$ at the end-effector:
\begin{equation}
\vc{\tau} = \vc{J}_x^T \vc{M}_x \ddot{\vc{x}} + \vc{g}(\vc{q}),
\end{equation}
where $\vc{q}$ is the current configuration of the robot in generalized coordinates, and $\vc{J}_x = \frac{\partial \vc{x}}{\partial \vc{q}}$ is the Jacobian evaluated at the end-effector. Since the push occurs at an equilibrium state, the effect of the Coriolis force is ignored, but the effect of gravity is included as $\vc{g}(\vc{q})$. The operational space mass matrix $\vc{M}_x$, using $\vc{M}$ as the original mass matrix in the generalized coordinates, is defined as:
\begin{equation}
    \vc{M}_x = (\vc{J}_x \vc{M}^{-1}\vc{J}_x^T)^{-1}.
\end{equation}

We compute the desired acceleration $\ddot{\vc{x}}$ using a simple PD feedback rule to match the desired linear position $\vc{x}_T^\text{des} \in \mathbf{R}^3$. We also command the robot to orient its end-effector to align with $\vc{x}_R^\text{des} \in \mathbf{R}^3$. For the derivative term, the end-effector is expected to match both the desired linear and angular velocities, $\dot{\vc{x}}^\text{des} \in \mathbf{R}^6$. Thus, the desired end-effector acceleration can be computed as: 
\begin{equation}
\label{eqn:desired-accel}
    \ddot{\vc{x}} = k_p\left(\begin{array}{c}\vc{x}_T^\text{des}-\vc{x}_T \\ \vc{x}_R^\text{des} \ominus \vc{x}_R \end{array}\right) + k_d(\dot{\vc{x}}^\text{des}- \dot{\vc{x}}),
\end{equation}
where $\vc{x} = [\vc{x}_T, \vc{x}_R] \in \mathbf{R}^6$ denotes the end-effector position $\vc{x}_T$ and orientation $\vc{x}_R$ in the world frame. The operator $\ominus$ indicates the difference between the two 3D orientations. The stiffness $k_p$ and damping coefficient $k_d$ are chosen manually ($k_d = 152$ and $k_p=3000$ in our experiments) to stabilize the controller.

During testing, we first query $\pi_\theta$ to obtain the optimal action $(a_1, a_2)$ for the current state. Then, we set the desired linear position in Equation \ref{eqn:desired-accel} to $\vc{x}_T^\text{des} = \vc{T}\vc{p}$, where $\vc{T}$ is the transformation that transforms a vector in the object frame to the world frame, and $\vc{p} \in \mathbf{R}^6$ corresponds to the coordinates in object frame of the point represented by $\vc{a}$. The desired orientation $\vc{x}_R^\text{des}$ is defined such that, when $\vc{x}_R = \vc{x}_R^\text{des}$, the end-effector is perpendicular to the surface $\vc{p}$ lies on.

Before the robot is in contact with the object, the desired velocity in Equation \ref{eqn:desired-accel} is set to zero, $\dot{\vc{x}}_T^\text{des} = \vc{0}$. When the end-effector is sufficiently close to the object and perpendicular to the surface of $\vc{p}$, we set the desired velocity to $\dot{\vc{x}}_T^\text{des} = \dot{\vc{p}}$ to exert the pushing force. We maintain this desired velocity for a fixed amount of time ($10$ time steps in our experiments). The object moves with the momentum imparted and eventually comes to rest due to the friction between the object and the surface. We execute actions in an open loop control and enforce safety limits on the robot torques.

\section{EXPERIMENTS}
We evaluate our method on a simulated KUKA robotic arm using the physics engine, DART \cite{Lee2018}, as well as on a real UR10 robot and compare our method to a predictor trained with trajectories from a random interaction policy.

\subsection{Network architecture and training}
The predictor network consists of fully connected and LSTM layers as shown in Figure \ref{fig:architecture}. The final softmax layer transforms the output of the last fully connected layer into the output mass distribution. Every fully connected layer except the last one is followed by the ReLU non-linearity. Similarly, we model the policy with two fully connected hidden layers containing $64$ neurons followed by an LSTM layer with $256$ units (Figure \ref{fig:architecture}). 


For the initial training of the predictor, we collect $16k$ and $32k$ episodes for the two-link and three-link chains respectively, by running a random policy that outputs a uniformly sampled action at each step until the end of the episode. The rollouts can be collected efficiently by running $16$ agents as parallel threads in $16$ environments initialized with different random seeds. We train the predictor for a total of $500k$ steps with a batch size of $16$, using stochastic gradient descent. The learning rate is initially set to $0.1$, and is halved every $166k$ iterations. This trained predictor is used as a part of the reward function for the subsequent policy training step. 

\begin{figure}[t]
\centering
\begin{subfigure}{0.25\textwidth}
\includegraphics[width=0.95\linewidth]{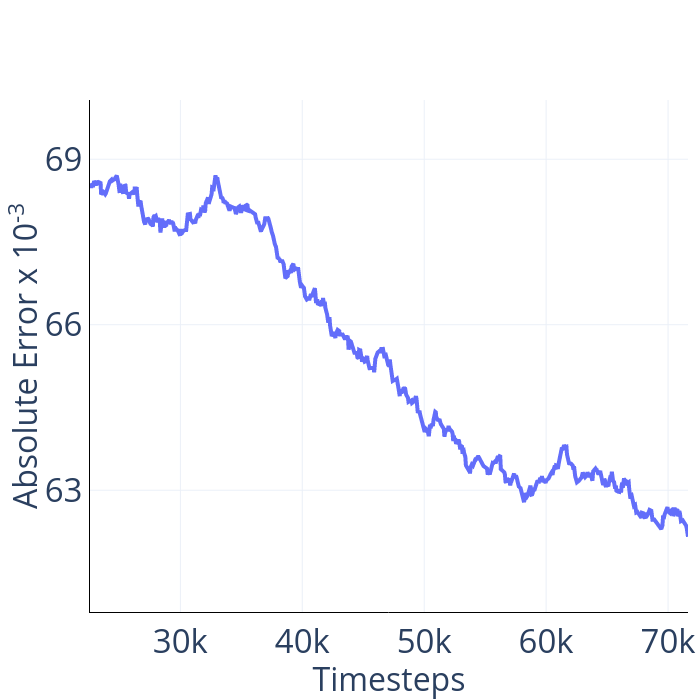}
\caption{}
\label{fig:ppo_error}
\end{subfigure}%
\begin{subfigure}{0.25\textwidth}
\includegraphics[width=0.95\linewidth]{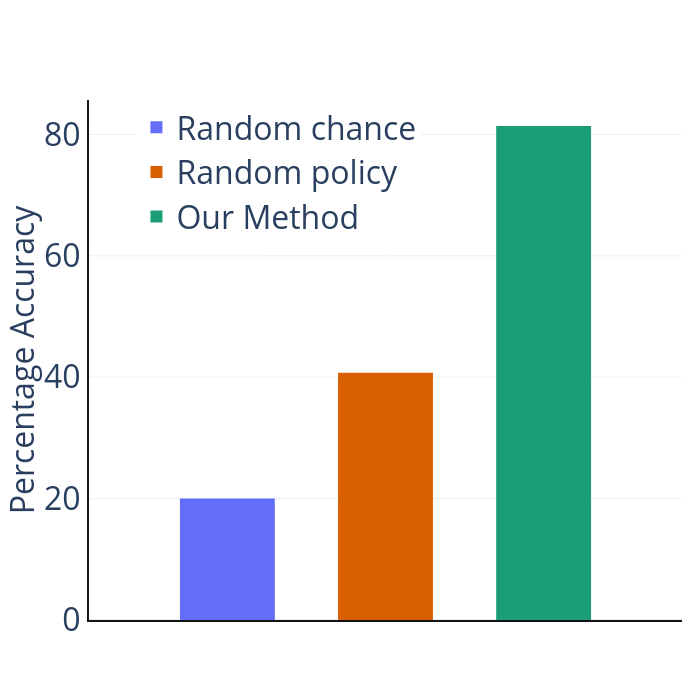}
\caption{}
\label{fig:real_accuracy}
\end{subfigure}%
\begin{subfigure}{0.25\textwidth}
\centering
\includegraphics[width=0.95\linewidth]{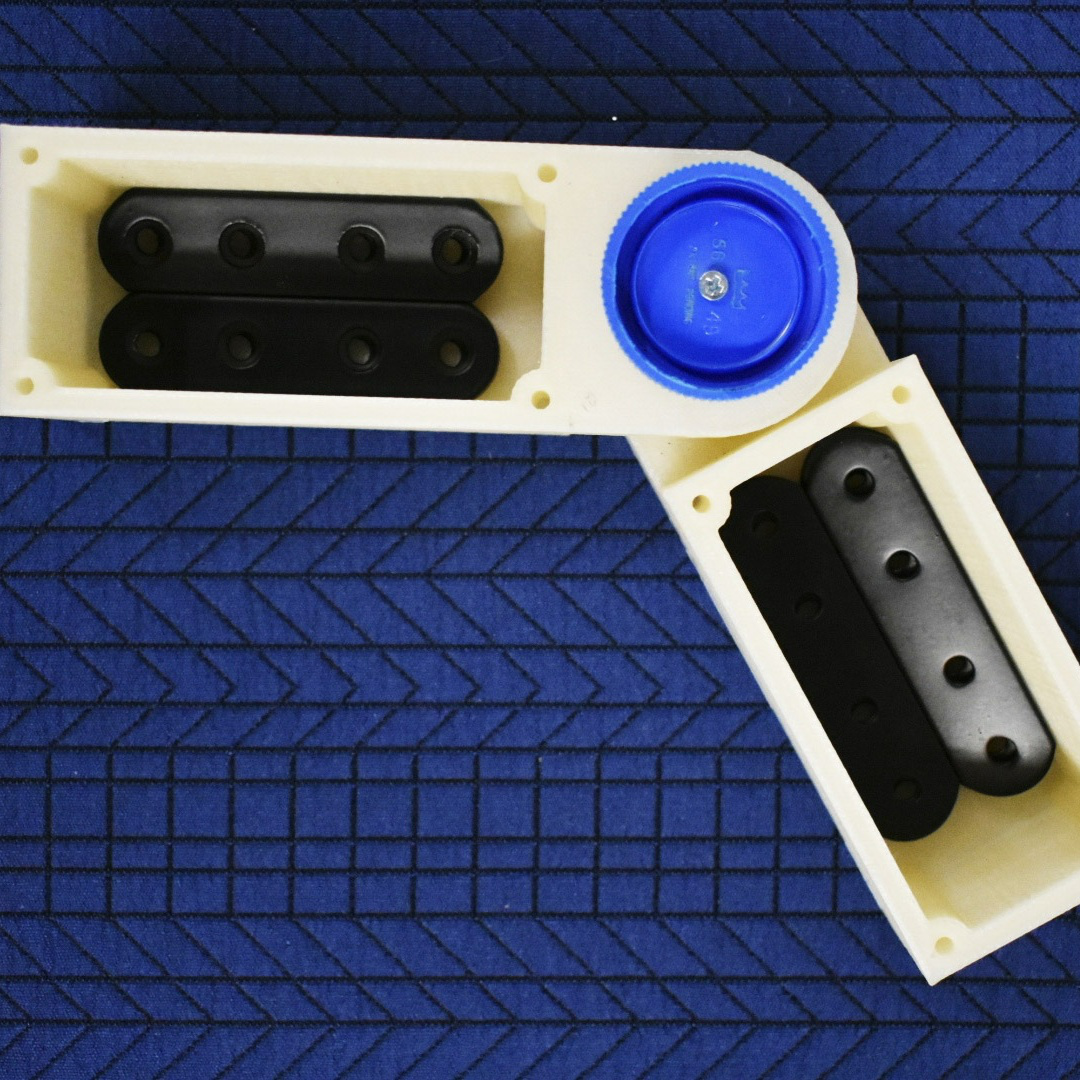}
\caption{}
\label{fig:mass_chamber}
\end{subfigure}%
\caption{(\textbf{\ref{fig:ppo_error}}): Prediction error (smoothed) in the first PPO meta-iteration for mass range $[0.1, 1.0]$. (\textbf{\ref{fig:real_accuracy}}): Accuracy on classification of two-link chains with $5$ different mass distributions in the real world . \textbf{(\ref{fig:mass_chamber})}: We vary the effective mass of each link by adding iron braces to the mass chamber }
\label{fig:bar_graph}
\end{figure}

The policy is trained using Proximal Policy Optimization (PPO) \cite{schulman2017proximal,stable-baselines} with an entropy loss term to promote exploration. We train the policy for $70k$ and $80k$ time steps, for two-link and three-link chains respectively, with a learning rate of $10^{-4}$. During this step the weights of the predictor network are frozen and only the policy weights are trained to maximize the expected reward and in-turn minimize the prediction error. Figure \ref{fig:ppo_error} shows the drop in error during one such policy meta-iteration. Once we train the policy, we freeze the policy network weights and fine-tune the predictor network for another $500k$ iterations using a fresh dataset of rollouts. It is essential to use a fresh dataset because the data distribution imposed by the neural network policy is different from the one imposed by the initial random policy. This process can be iterated multiple times. In our experiments, we converge after six meta-iterations of policy training and predictor fine-tuning. As the learning progresses, the improvement from each meta-iteration decreases and eventually becomes unnoticeable. We train the predictor at the last meta-iteration with a deterministic policy that leads to a further drop in error as seen in Figure \ref{fig:meta_iter_2link} and Figure \ref{fig:meta_iter_3link}. We call this the Trained Policy Predictor plus(TP+) model.

We train three baseline models to compare the performance of our model against:  (i) Random Policy Predictor (RP) trained with $32k$ episodes for two-link chain and $64k$ episodes for three-link chain. (ii) Random Policy Predictor plus (RP+) trained with $64k$ episodes for the two-link chain and $128k$ episodes for the three-link chain. (iii) Trained Policy predictor (TP) trained with a total of $64k$ episodes for the two-link chain and $128k$ episodes for the three-link chain but with one intermediate policy meta-iteration.
The Random Policy (RP) predictor models are similar to \cite{24} where the action taken is independent of the configuration of the object.


\subsection{Evaluation in simulation}
We import a model of the KUKA KR $5$ Sixx R$650$ Robotic arm in DART and an articulated chain of rigid cuboid segments connected by revolute joints. We experiment with two types of objects: 2-link chains and 3-link chains, with the link masses uniformly sampled from three different mass ranges: $[0.1, 1]$, $[1, 10]$ and $[0.1, 10]$. A wider mass range is more challenging because the set of informative pushing forces needs to be more diverse. When the robot uses a large force to push a light object, the object might rapidly rotate over many cycles with multiple self-collisions between the links. This information is lost in the training set because the observation only records the equilibrium state of the object, rather than the entire trajectory. Consequently, the training set contains examples that result in many-to-one mapping between input and output spaces.


In addition to varying the mass, we also randomize the friction coefficient from a range of $[0.5, 1]$ and the length of each segment from $[0.1, 0.15]$. We start the episode with a random configuration of the articulated object and add Gaussian noise with zero mean and $0.01$ standard deviation to the observations and actions in the environment. The primary motivation to introduce noise into our simulations is to prevent the networks from over-fitting to the simulator's deterministic dynamics and to ensure that the networks trained in simulation can be reliable transferred to the real world.


We summarize the performance of the models in Table \ref{tab:Ablation}. We find that having a policy that strategically pushes with the intention of minimizing prediction error outperforms the random policy across all cases. This observation falls in line with evidence from neuroscience community on the importance of active inference in learning a model of the world ~\cite{friston2017active, bohg2017interactive}.  We also compare the improvement a single policy meta-iteration (TP) offers to the performance of the predictor. Even though predictor models RP+ and TP are trained with the same amount of training data, TP achieves a lower percentage error across all cases, as shown in Table \ref{tab:Ablation}. Further, performance improvement between RP and RP+ is minuscule in spite of doubling the dataset size. We observe that the improvement our policy offers for wider and more difficult mass ranges is more significant than for smaller ranges. We also note that alternating between supervised training and reinforcement learning is quite stable and leads to a consistent drop in percentage error as show in Figure \ref{fig:meta_iter_2link} and \ref{fig:meta_iter_3link}.
\begin{table}[t]
\centering
\begin{tabular}{ |c|c|c|c|c| } 
 \hline
 Link Type & RP & RP+ & TP (Ours) & TP+ (Ours) \\
 \hline
 2-Link $[0.1, 1.0]$ & 15.26 & 14.8 & 10.39 & \textbf{6.07}\\ 
 2-Link $[1.0, 10]$ & 17.12 & 17.4 & 14.03 & \textbf{8.26}\\
 2-Link $[0.1, 10]$ & 26.68 & 26.78 & 21.27 & \textbf{12.03}\\
 3-Link $[0.1,1.0]$ & 29.08 & 28.88 & 26.89 & \textbf{12.02}\\
 3-Link $[1.0, 10]$ & 30.88 & 30.84 & 29.75 & \textbf{15.99}\\
 3-Link $[0.1, 10]$ & 54.28 & 53.59 & 50.49 & \textbf{23.93}\\
 \hline

\end{tabular}
 \caption{\label{tab:Ablation} Comparison of our method to the baselines. RP and RP+ predictors are trained with a random interaction policy with different dataset sizes. TP predictor uses the policy obtained after one policy meta-iteration while TP+ uses the policy obtained after 6 meta-iterations.}
\end{table}
\subsection{Evaluation on real hardware}
In this section, we show that our method that is trained in simulation can be transferred to the real world by fine-tuning on a small amount of real-world data. Unlike simulation, moving the base of the real robot around to follow the object can be unrealistic. With a stationary robot, the mass ranges that can be explored become limited as the object can be pushed out of the robot workspace easily with one wrong action.
We choose a UR10 robot which has a significantly larger workspace than the KUKA robot to conduct our experiments. We make the robot interact with articulated 3D printed chains of different mass distributions. The $3$D printed links have a designated ``mass chamber'', shown in Figure \ref{fig:mass_chamber}, where different weights can be added to vary the mass distribution of the object as necessary. We connect these links with ball-bearings to create a near friction-free revolute joint. A Realsense D435 depth camera is utilized to capture RGBD images of the scene consisting of the table, the robot and the articulated object. We calibrate the entire setup by using QR codes placed on various spots on the table with a ROS wrapper for the alvar AR tag tracking library. Further, we place QR codes on each link and track the pose of the articulated chain after every robot interaction. 

To manipulate the chain, the robot has to hit it at precise locations as instructed by the policy. We design an end-effector with a wooden dowel and attach it to the robot with a 3D printed mount in our experiments. To ensure that the contact with the object is at a precise point on the object, we tilt the dowel by an angle of approximately $30^{\circ}$. After every interaction, the robot resets to its rest state so that the camera can get an occlusion free image of the object. The agent registers the new location of the object and performs the next interaction step as instructed by the policy.

The real-world experiments are conducted using a two-link chain with a joint range of $[-\pi/2,\pi/2]$. We first use the uniform policy to collect $30$ episodes each ($150$ in total) of the robot interacting with $5$ different mass distributions. The two-link objects are visually identical and can be one of the following weight pairs which span the range of mass distributions possible: $[0.16,0.16]$, $[0.064,0.192]$, $[0.064, 0.256]$, $[0.192, 0.064]$ and $[0.256,0.064]$. We simulate the robot in PyBullet \cite{coumans2019} and train a predictor on the mass range $[0.01,0.3]$. The features learnt by the predictor are fed to a small two-layer fully connected network $[64-Relu-32-Relu-5]$, trained to classify the objects into one of the $5$ classes by using a cross-entropy loss. We find that the Random Policy Predictor reaches an accuracy of $40.74\%$, with random chance being $20\%$. We then collect another $150$ episodes with our learned policy and train the classification network using the features from the predictor of the last meta iteration. Our method outperforms the baseline by a large margin achieving an accuracy of $81.4\%$ as shown in Figure \ref{fig:real_accuracy}. We notice that the learned policy tries to push the object such that there is movement about the revolute joint so that inference about individual links is possible. Intuitively this makes sense, as in cases where the joint is locked down, the body behaves like a rigid object and it becomes hard to decouple properties about individual links in the presence of real-world sensing and actuation noise. Although we need significant amount of training data in simulation, the real world experiments show that it is possible to fine-tune the models trained in simulation with reasonable amount of real world data.


\section{DISCUSSION}
We present an agent that learns to predict the mass distribution of articulated chains by strategically selecting few pushes that extract maximum information. We propose a training procedure that alternates between training a predictor with supervised learning, and a policy with reinforcement learning. We test our hypothesis in simulation by pushing 2-link and 3-link articulated chains with a robotic arm. Our experiments show that our approach consistently outperforms the baseline method. 
We also use a real UR10 robot to test out our algorithm on a 2-link chain in the real world.
 
One drawback of our method is that the action space is tightly coupled to the geometry of the object we are trying to manipulate. This requires us to train a separate network for every object type. However, interaction strategies learnt by interacting with one object often transfer to similar object types. In future, we want to design architectures that can be applied across a family of object types by potentially leveraging an image based observation space.
\subsubsection*{Acknowledgments}
We would like to thank Dr. Matthew Gombolay for letting us use the UR10 robot and Harish Haresamudram for proofreading the paper.
\bibliography{IEEEabrv,draft1}

\begin{thebibliography}{41}
\providecommand{\natexlab}[1]{#1}
\providecommand{\url}[1]{\texttt{#1}}
\expandafter\ifx\csname urlstyle\endcsname\relax
  \providecommand{\doi}[1]{doi: #1}\else
  \providecommand{\doi}{doi: \begingroup \urlstyle{rm}\Url}\fi

\bibitem[Spelke and Kinzler(2007)]{Spelke2007}
E.~S. Spelke and K.~D. Kinzler.
\newblock {Core knowledge}.
\newblock \emph{Developmental Science}, 10\penalty0 (1):\penalty0 89--96, jan
  2007.
\newblock ISSN 1363755X.
\newblock \doi{10.1111/j.1467-7687.2007.00569.x}.

\bibitem[Hamrick et~al.()Hamrick, Battaglia, and
  Tenenbaum]{Hamrick_internalphysics}
J.~Hamrick, P.~Battaglia, and J.~B. Tenenbaum.
\newblock Internal physics models guide probabilistic judgments about object
  dynamics.

\bibitem[Baldwin et~al.(1993)Baldwin, Markman, and Melartin]{10.2307/1131213}
D.~A. Baldwin, E.~M. Markman, and R.~L. Melartin.
\newblock Infants' ability to draw inferences about nonobvious object
  properties: Evidence from exploratory play.
\newblock \emph{Child Development}, 64\penalty0 (3):\penalty0 711--728, 1993.
\newblock ISSN 00093920, 14678624.

\bibitem[Muentener et~al.(2018)Muentener, Herrig, and
  Schulz]{10.3389/fpsyg.2018.00635}
P.~Muentener, E.~Herrig, and L.~Schulz.
\newblock The efficiency of infants' exploratory play is related to longer-term
  cognitive development.
\newblock \emph{Frontiers in Psychology}, 9:\penalty0 635, 2018.
\newblock ISSN 1664-1078.

\bibitem[Wu et~al.(2015)Wu, Yildirim, Lim, Freeman, and Tenenbaum]{14}
J.~Wu, I.~Yildirim, J.~J. Lim, B.~Freeman, and J.~Tenenbaum.
\newblock Galileo: Perceiving physical object properties by integrating a
  physics engine with deep learning.
\newblock In \emph{Advances in Neural Information Processing Systems 28}, pages
  127--135. 2015.

\bibitem[Zheng et~al.(2018)Zheng, Luo, Wu, and Tenenbaum]{12}
D.~Zheng, V.~Luo, J.~Wu, and J.~B. Tenenbaum.
\newblock Unsupervised learning of latent physical properties using
  perception-prediction networks.
\newblock \emph{arXiv preprint arXiv:1807.09244}, 2018.

\bibitem[Xu et~al.(2019)Xu, Wu, Zeng, Tenenbaum, and Song]{24}
Z.~Xu, J.~Wu, A.~Zeng, J.~B. Tenenbaum, and S.~Song.
\newblock Densephysnet: Learning dense physical object representations via
  multi-step dynamic interactions.
\newblock In \emph{RSS}, 2019.

\bibitem[Kubricht et~al.(2017)Kubricht, Holyoak, and Lu]{46}
J.~R. Kubricht, K.~J. Holyoak, and H.~Lu.
\newblock Intuitive physics: Current research and controversies.
\newblock \emph{Trends in cognitive sciences}, 21\penalty0 (10):\penalty0
  749--759, 2017.

\bibitem[Hamrick et~al.(2016)Hamrick, Battaglia, Griffiths, and Tenenbaum]{47}
J.~B. Hamrick, P.~W. Battaglia, T.~L. Griffiths, and J.~B. Tenenbaum.
\newblock Inferring mass in complex scenes by mental simulation.
\newblock \emph{Cognition}, 157:\penalty0 61--76, 2016.

\bibitem[Nguyen-Tuong and Peters(2011)]{48}
D.~Nguyen-Tuong and J.~Peters.
\newblock Model learning for robot control: a survey.
\newblock \emph{Cognitive processing}, 12\penalty0 (4):\penalty0 319--340,
  2011.

\bibitem[Battaglia et~al.(2016)Battaglia, Pascanu, Lai, Rezende, et~al.]{2}
P.~Battaglia, R.~Pascanu, M.~Lai, D.~J. Rezende, et~al.
\newblock Interaction networks for learning about objects, relations and
  physics.
\newblock In \emph{NIPS}, pages 4509--4517, 2016.

\bibitem[Chang et~al.(2016)Chang, Ullman, Torralba, and Tenenbaum]{5}
M.~B. Chang, T.~Ullman, A.~Torralba, and J.~B. Tenenbaum.
\newblock A compositional object-based approach to learning physical dynamics.
\newblock \emph{arXiv preprint arXiv:1612.00341}, 2016.

\bibitem[Purushwalkam et~al.(2019)Purushwalkam, Gupta, Kaufman, and Russell]{6}
S.~Purushwalkam, A.~Gupta, D.~Kaufman, and B.~Russell.
\newblock Bounce and learn: Modeling scene dynamics with real-world bounces.
\newblock In \emph{ICLR}, 2019.

\bibitem[Ye et~al.(2018)Ye, Wang, Davidson, and Gupta]{8}
T.~Ye, X.~Wang, J.~Davidson, and A.~Gupta.
\newblock Interpretable intuitive physics model.
\newblock In \emph{ECCV}, 2018.

\bibitem[Ajay et~al.(2018)Ajay, Wu, Fazeli, Bauza, Kaelbling, Tenenbaum, and
  Rodriguez]{13}
A.~Ajay, J.~Wu, N.~Fazeli, M.~Bauza, L.~P. Kaelbling, J.~B. Tenenbaum, and
  A.~Rodriguez.
\newblock Augmenting physical simulators with stochastic neural networks: Case
  study of planar pushing and bouncing.
\newblock \emph{2018 IEEE/RSJ IROS}, Oct 2018.
\newblock \doi{10.1109/iros.2018.8593995}.

\bibitem[Wu et~al.(2017)Wu, Lu, Kohli, Freeman, and Tenenbaum]{27}
J.~Wu, E.~Lu, P.~Kohli, B.~Freeman, and J.~Tenenbaum.
\newblock Learning to see physics via visual de-animation.
\newblock In \emph{Advances in Neural Information Processing Systems}, pages
  153--164, 2017.

\bibitem[Ehrhardt et~al.(2019)Ehrhardt, Monszpart, Mitra, and Vedaldi]{28}
S.~Ehrhardt, A.~Monszpart, N.~J. Mitra, and A.~Vedaldi.
\newblock Taking visual motion prediction to new heightfields.
\newblock \emph{Computer Vision and Image Understanding}, 181:\penalty0 14--25,
  2019.

\bibitem[Mrowca et~al.(2018)Mrowca, Zhuang, Wang, Haber, Fei-Fei, Tenenbaum,
  and Yamins]{40}
D.~Mrowca, C.~Zhuang, E.~Wang, N.~Haber, L.~F. Fei-Fei, J.~Tenenbaum, and D.~L.
  Yamins.
\newblock Flexible neural representation for physics prediction.
\newblock In \emph{Advances in Neural Information Processing Systems}, pages
  8799--8810, 2018.

\bibitem[Battaglia et~al.(2013)Battaglia, Hamrick, and Tenenbaum]{4}
P.~W. Battaglia, J.~B. Hamrick, and J.~B. Tenenbaum.
\newblock Simulation as an engine of physical scene understanding.
\newblock \emph{Proceedings of the National Academy of Sciences}, 110\penalty0
  (45):\penalty0 18327--18332, 2013.
\newblock ISSN 0027-8424.
\newblock \doi{10.1073/pnas.1306572110}.

\bibitem[Li et~al.(2019)Li, Leonardis, Bohg, and Fritz]{22}
W.~Li, A.~Leonardis, J.~Bohg, and M.~Fritz.
\newblock Learning manipulation under physics constraints with visual
  perception.
\newblock \emph{arXiv preprint arXiv:1904.09860}, 2019.

\bibitem[Li et~al.(2017)Li, Leonardis, and Fritz]{41}
W.~Li, A.~Leonardis, and M.~Fritz.
\newblock Visual stability prediction and its application to manipulation.
\newblock In \emph{2017 AAAI Spring Symposium Series}, 2017.

\bibitem[Fragkiadaki et~al.(2015)Fragkiadaki, Agrawal, Levine, and Malik]{21}
K.~Fragkiadaki, P.~Agrawal, S.~Levine, and J.~Malik.
\newblock Learning visual predictive models of physics for playing billiards.
\newblock \emph{arXiv preprint arXiv:1511.07404}, 2015.

\bibitem[Finn and Levine(2017)]{32}
C.~Finn and S.~Levine.
\newblock Deep visual foresight for planning robot motion.
\newblock In \emph{2017 IEEE ICRA}, pages 2786--2793. IEEE, 2017.

\bibitem[Agrawal et~al.(2016)Agrawal, Nair, Abbeel, Malik, and Levine]{20}
P.~Agrawal, A.~V. Nair, P.~Abbeel, J.~Malik, and S.~Levine.
\newblock Learning to poke by poking: Experiential learning of intuitive
  physics.
\newblock In \emph{NIPS}, pages 5074--5082, 2016.

\bibitem[Yu et~al.(2017)Yu, Tan, Liu, and Turk]{33}
W.~Yu, J.~Tan, C.~K. Liu, and G.~Turk.
\newblock Preparing for the unknown: Learning a universal policy with online
  system identification.
\newblock \emph{arXiv preprint arXiv:1702.02453}, 2017.

\bibitem[Zhou et~al.(2019)Zhou, Pinto, and Gupta]{34}
W.~Zhou, L.~Pinto, and A.~Gupta.
\newblock Environment probing interaction policies.
\newblock In \emph{ICLR}, 2019.

\bibitem[Yang et~al.(2019)Yang, Petersen, Zha, and Faissol]{yang2019single}
J.~Yang, B.~Petersen, H.~Zha, and D.~Faissol.
\newblock Single episode policy transfer in reinforcement learning.
\newblock \emph{arXiv preprint arXiv:1910.07719}, 2019.

\bibitem[Sturm et~al.(2009)Sturm, Pradeep, Stachniss, Plagemann, Konolige, and
  Burgard]{18}
J.~Sturm, V.~Pradeep, C.~Stachniss, C.~Plagemann, K.~Konolige, and W.~Burgard.
\newblock Learning kinematic models for articulated objects.
\newblock In \emph{IJCAI}, 2009.

\bibitem[Venkataraman et~al.(2018)Venkataraman, Griffin, and Corso]{23}
A.~Venkataraman, B.~Griffin, and J.~J. Corso.
\newblock Learning kinematic descriptions using spare: Simulated and physical
  articulated extendable dataset.
\newblock \emph{arXiv preprint arXiv:1803.11147}, 2018.

\bibitem[{Brock} et~al.(2009){Brock}, {Trinkle}, and {Ramos}]{15}
O.~{Brock}, J.~{Trinkle}, and F.~{Ramos}.
\newblock \emph{Learning to Manipulate Articulated Objects in Unstructured
  Environments Using a Grounded Relational Representation}.
\newblock MITP, 2009.
\newblock ISBN 9780262258623.

\bibitem[Katz et~al.(2013)Katz, Kazemi, Bagnell, and Stentz]{17}
D.~Katz, M.~Kazemi, J.~A. Bagnell, and A.~Stentz.
\newblock Interactive segmentation, tracking, and kinematic modeling of unknown
  3d articulated objects.
\newblock \emph{2013 IEEE ICRA}, pages 5003--5010, 2013.

\bibitem[Katz et~al.(2014)Katz, Orthey, and Brock]{19}
D.~Katz, A.~Orthey, and O.~Brock.
\newblock Interactive perception of articulated objects.
\newblock In \emph{Experimental Robotics}, pages 301--315. Springer, 2014.

\bibitem[Bohg et~al.(2017)Bohg, Hausman, Sankaran, Brock, Kragic, Schaal, and
  Sukhatme]{bohg2017interactive}
J.~Bohg, K.~Hausman, B.~Sankaran, O.~Brock, D.~Kragic, S.~Schaal, and G.~S.
  Sukhatme.
\newblock Interactive perception: Leveraging action in perception and
  perception in action.
\newblock \emph{IEEE Transactions on Robotics}, 33\penalty0 (6):\penalty0
  1273--1291, 2017.

\bibitem[{Yong Yu} et~al.(2005){Yong Yu}, {Arima}, and {Tsujio}]{26}
{Yong Yu}, T.~{Arima}, and S.~{Tsujio}.
\newblock Estimation of object inertia parameters on robot pushing operation.
\newblock In \emph{Proceedings of the 2005 IEEE ICRA}, pages 1657--1662, April
  2005.

\bibitem[Wierstra et~al.(2007)Wierstra, Foerster, Peters, and
  Schmidhuber]{wierstra2007solving}
D.~Wierstra, A.~Foerster, J.~Peters, and J.~Schmidhuber.
\newblock Solving deep memory pomdps with recurrent policy gradients.
\newblock In \emph{International Conference on Artificial Neural Networks},
  pages 697--706. Springer, 2007.

\bibitem[Schulman et~al.(2017)Schulman, Wolski, Dhariwal, Radford, and
  Klimov]{schulman2017proximal}
J.~Schulman, F.~Wolski, P.~Dhariwal, A.~Radford, and O.~Klimov.
\newblock Proximal policy optimization algorithms.
\newblock \emph{arXiv preprint arXiv:1707.06347}, 2017.

\bibitem[Khatib(1987)]{khatib1987unified}
O.~Khatib.
\newblock A unified approach for motion and force control of robot
  manipulators: The operational space formulation.
\newblock \emph{IEEE Journal on Robotics and Automation}, 3\penalty0
  (1):\penalty0 43--53, 1987.

\bibitem[Lee et~al.(2018)Lee, Grey, Ha, Kunz, Jain, Ye, Srinivasa, Stilman, and
  Liu]{Lee2018}
J.~Lee, M.~X. Grey, S.~Ha, T.~Kunz, S.~Jain, Y.~Ye, S.~S. Srinivasa,
  M.~Stilman, and C.~K. Liu.
\newblock {DART}: Dynamic animation and robotics toolkit.
\newblock \emph{The Journal of Open Source Software}, 3\penalty0 (22):\penalty0
  500, Feb 2018.
\newblock \doi{10.21105/joss.00500}.
\newblock URL \url{https://doi.org/10.21105/joss.00500}.

\bibitem[Hill et~al.(2018)Hill, Raffin, Ernestus, Traore, Dhariwal, Hesse,
  Klimov, Nichol, Plappert, Radford, Schulman, Sidor, and Wu]{stable-baselines}
A.~Hill, A.~Raffin, M.~Ernestus, R.~Traore, P.~Dhariwal, C.~Hesse, O.~Klimov,
  A.~Nichol, M.~Plappert, A.~Radford, J.~Schulman, S.~Sidor, and Y.~Wu.
\newblock Stable baselines, 2018.

\bibitem[Friston et~al.(2017)Friston, FitzGerald, Rigoli, Schwartenbeck, and
  Pezzulo]{friston2017active}
K.~Friston, T.~FitzGerald, F.~Rigoli, P.~Schwartenbeck, and G.~Pezzulo.
\newblock Active inference: a process theory.
\newblock \emph{Neural computation}, 29\penalty0 (1):\penalty0 1--49, 2017.

\bibitem[Coumans and Bai(2016--2019)]{coumans2019}
E.~Coumans and Y.~Bai.
\newblock Pybullet, a python module for physics simulation for games, robotics
  and machine learning.
\newblock \url{http://pybullet.org}, 2016--2019.

\end{thebibliography}
\appendix
\section{Are all forces equally good?}\label{proof:1}
Consider an arbitrary articulated object that has a mass matrix $\mathcal{M}(q)$. We use $q \in \rm I\!{R}^N$ to represent the configuration of the object in generalized coordinates.
\begin{equation*}
\begin{aligned}
\mathcal{M}(q) \ddot{q} + C(q, \dot{q})& = Q, \\
\end{aligned}
\end{equation*}
where $C$ is the coriolis force and $Q$ is the external force both expressed in the generalized coordinates.

Let $J_k \in \rm I\!{R}^{6\times N}$ be the Jacobian and $I_k \in \rm I\!{R}^{3\times 3}$ be the inertia matrix of the rigid body $k$. $J_k$ can further be split into the linear and angular constituent matrices  $J_v \in \rm I\!{R}^{3\times N}$ and $J_\omega \in \rm I\!{R}^{3\times N}$ respectively i.e. $J_k = \begin{bmatrix}
J_v \\
J_\omega
\end{bmatrix}$. By definition,
\begin{equation*}
\begin{aligned}
\mathcal{M}(q) & = \sum_{k} J^T_{k}(q)
\begin{bmatrix}
m_kI & 0 \\
0 & m_kI_k
\end{bmatrix} J_k(q) \\
& = \sum_{k} m_k (J_{v_k}^T J_{v_k} + J_{\omega_{k}}^T I_k J{\omega_{k}}). \\
\end{aligned}
\end{equation*}

Assuming $\dot{q} =0 \implies C(q,\dot{q}) = 0$, the equation of motion can be expressed as
\begin{equation*}
    \sum_{k} m_k A_k \ddot{q} = Q,
\end{equation*}
where $A_k = J_{v_k}^T J_{v_k} + J_{\omega_{k}}^T I_k J{\omega_{k}}$.

If $N > 6$, $A_k \in \rm I\!{R}^{N \times N}$ is not full-rank by definition. Thus there exists a null space for each $A_k$. If $\ddot{q}$ lies in the null-space of any $A_k$ we cannot uniquely determine the value of $m_k$. If $N \leq 6$, we still cannot guarantee $A_k$ to be full-rank because if the object is in a kinematic singularity state, $J_k(q)$ will lose rank.

\end{document}